\documentclass[%
 reprint,
 amsmath,amssymb,
 aps, 
 pre,
 showkeys
]{revtex4-1}

\usepackage{graphicx}
\usepackage{dcolumn}
\usepackage{bm}

\usepackage{my-style}
\graphicspath{{figs/}}
\usepackage{algpseudocode, algorithm}
\usepackage{multirow}

\allowdisplaybreaks[4]

\begin{document}

\preprint{APS/123-QED}

\title{Spatial Monte Carlo Integration with Annealed Importance Sampling}

\author{Muneki Yasuda}  
\email{muneki@yz.yamagata-u.ac.jp}
\affiliation{%
 Graduate School of Science and Engineering, Yamagata University, Japan.
}%
\author{Kaiji Sekimoto}  
\affiliation{%
 Graduate School of Science and Engineering, Yamagata University, Japan.
}%


\begin{abstract}
Evaluating expectations on an Ising model (or Boltzmann machine) is essential for various applications, including statistical machine learning. 
However, in general, the evaluation is computationally difficult because it involves intractable multiple summations or integrations; 
therefore, it requires approximation. 
Monte Carlo integration (MCI) is a well-known approximation method; 
a more effective MCI-like approximation method was proposed recently, called spatial Monte Carlo integration (SMCI). 
However, the estimations obtained using SMCI (and MCI) exhibit a low accuracy in Ising models under a low temperature owing to degradation of the sampling quality.  
Annealed importance sampling (AIS) is a type of importance sampling based on Markov chain Monte Carlo methods 
that can suppress performance degradation in low-temperature regions with the force of importance weights. 
In this study, a new method is proposed to evaluate the expectations on Ising models combining AIS and SMCI. 
The proposed method performs efficiently in both high- and low-temperature regions, which is demonstrated theoretically and numerically.
\end{abstract}

\pacs{Valid PACS appear here}
\keywords{Boltzmann machine, inference, spatial Monte Carlo integration, annealed importance sampling}
\maketitle


\section{Introduction}

An Ising model, also known as a Boltzmann machine~\cite{PBM1985,Roudi2009}, 
is one of the most important models in not only statistical physics but also other various fields, such as machine learning and optimization. 
For example, in the field of machine learning, the Boltzmann machine and its variants, 
such as restricted Boltzmann machine~\cite{RBM1986,CD2002,GBRBM2011,DRBM2012,CDRBM2019,GSRBM2020} 
and deep Boltzmann machine~\cite{DBM2009,DBM2012,DBM2013,DeepGBRBM2013}, have been actively studied. 
Evaluating expectations on Ising models is essential for such applications. 
However, the evaluation is generally computationally difficult because it involves intractable multiple summations or integrations.
This study aims to propose an effective approximation for the evaluation. 

Monte Carlo integration (MCI) is the most familiar sampling approximation, 
in which a target expectation on an Ising model is approximated by the sample average over a sample set;  
the sampling points are generated using Markov chain Monte Carlo (MCMC) methods on the Ising model. 
Recently, a more effective MCI-like method, called spatial Monte Carlo integration (SMCI), 
was proposed as an extension of MCI~\cite{SMCI2015,SMCI2020} (see section \ref{sec:SMCI}). 
It has been proved that SMCI is statistically more accurate than MCI.
The performances of MCI and SMCI are directly dependent on the sampling quality. 
The estimations obtained using these methods are of substandard quality when the sample set has an unexpected bias.
Gibbs sampling~\cite{Geman&Geman1984} has been widely used as a sampling method. 
However, Gibbs sampling tends to fail when the distribution structure is complicated, e.g., when there are several isolated modes; 
this is known as the slow relaxation problem.  
The influence of this problem is particularly prominent in Ising models under low temperatures (see section \ref{sec:experiment_AIS_vs_SMCI}).
To resolve this problem, sophisticated sampling methods, such as parallel tempering (PT) (or replica exchange MCMC)~\cite{PT1986,ExMCMC1996}, have been proposed. 
Nevertheless, Gibbs sampling is still preferred in terms of cost and implementation. 

Annealed importance sampling (AIS) is a type of importance sampling based on MCMC with simulated annealing~\cite{AIS2001} (see section \ref{sec:AIS}). 
In AIS, a sequential sampling (or ancestral sampling) from a tractable initial distribution to the target distribution is executed, 
in which the transitions between the distributions are executed using, for example, Gibbs sampling.  
AIS can suppress the performance degradation of the sampling approximation in Ising models under low temperatures (see section \ref{sec:experiment_AIS_vs_SMCI}). 
In this study, a new sampling approximation is proposed for Ising models by combining AIS and SMCI, 
which can provide accurate approximations in both high- and low-temperature regions. 
The proposed method is based on the usual Gibbs sampling.

The remainder of this paper is organized as follows. 
The Ising model used in this study is described in section \ref{sec:PBM}. 
SMCI and AIS are explained in section \ref{sec:sampling_approximation};  
this section also examines the results of numerical experiments, 
in which the influence of the slow relaxation problem of Gibbs sampling was observed using MCI and SMCI. 
The proposed method, i.e., AIS-based SMCI, is described in section \ref{sec:AIS+SMCI}, 
and the validation of the proposed method through numerical experiments is presented in section \ref{sec:experiment_AIS+SMCI}, 
in which the computational efficiency of the proposed method and a comparison with PT are also discussed.
Finally, the summary along with the future scope of the study are presented in section \ref{sec:summary}.

\section{Ising model}
\label{sec:PBM}

Consider an undirected graph $G(\mcal{V}, \mcal{E})$, where $\mcal{V} =\{ 1,2,\ldots, n\}$ is the set of vertices, 
and $\mcal{E}$ is the set of undirected edges in which the edge between vertices $i$ and $j$ is labeled as $(i,j)$. 
Because the edges have no direction, $(i,j)$ and $(j,i)$ indicate the same edge.
On this undirected graph, consider an energy function (or a Hamiltonian) with a quadratic form, as follows: 
\begin{align}
E(\bm{x}):= -\sum_{i \in \mcal{V}} h_i x_i - \sum_{(i,j) \in \mcal{E}} J_{i,j} x_i x_j,
\label{eqn:Hamiltonian}
\end{align}
where $\bm{x} := \{ x_i \in \{-1,+1\} \mid i \in \mcal{V}\}$ denotes the random (Ising) variables assigned to the corresponding nodes. 
Here, $h_i$ is the bias (or local field) on vertex $i$ and $J_{i,j}$ is the interaction between $i$ and $j$; 
the interactions are symmetric with respect to their indices, i.e., $J_{i,j} = J_{j,i}$.
Using the energy function, an Ising model is defined as
\begin{align}
P(\bm{x} \mid \beta):=\frac{1}{Z(\beta)} \exp\big(- \beta E(\bm{x}) \big),
\label{eqn:PBM}
\end{align}
where $\beta \geq 0$ is the inverse temperature and $Z(\beta)$ is the partition function defined by 
\begin{align}
Z(\beta) := \sum_{\bm{x}}\exp\big(- \beta E(\bm{x}) \big),
\end{align}
where $\sum_{\bm{x}}$ is the summation over all possible realizations of $\bm{x}$. 

The main aim of this study is to investigate an effective approximation method for the expectation of $f(\bm{x})$:
\begin{align}
\ave{f(\bm{x})}_{\beta} := \sum_{\bm{x}}f(\bm{x})P(\bm{x} \mid \beta).
\label{eqn:expectation-f}
\end{align}
The evaluation of this expectation is computationally infeasible because its general computational cost is $O(2^n)$.

\section{Sampling Approximations}
\label{sec:sampling_approximation}

MCI is one of the most frequently used methods for approximating equation (\ref{eqn:expectation-f}), in which 
the expectation is approximated by
\begin{align}
\ave{f(\bm{x})}_{\beta} \approx \frac{1}{N}\sum_{\mu=1}^N f(\mbf{s}_{\mu}), 
\label{eqn:expectation-f_MCI}
\end{align}
where $\mathbb{S} := \{ \mbf{s}_{\mu} \in \{-1,+1\}^n \mid \mu = 1,2,\ldots,N\}$ is the (i.i.d.) sample set drawn from $P(\bm{x} \mid \beta)$.
In this section, SMCI~\cite{SMCI2015,SMCI2020} and AIS~\cite{AIS2001}, which are effective approximation methods, are briefly described; 
subsequently, their performances are compared through numerical experiments.  

\subsection{Spatial Monte Carlo integration}
\label{sec:SMCI}

\begin{figure}[tb]
\centering
\includegraphics[height=3.5cm]{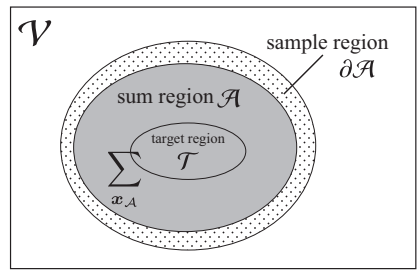}
\caption{Illustration of the target, sum, and sample regions of SMCI.}
\label{fig:region_sheme_SMCI}
\end{figure}

Here, the approximation of the expectation of $f(\bm{x}_{\mcal{T}})$ is considered, where $\mcal{T}$ is a (connected) subregion of $\mcal{V}$ and 
$\bm{x}_{\mcal{T}} := \{x_i \mid i \in \mcal{T} \subseteq \mcal{V}\}$ denotes the variables in $\mcal{T}$.
For subregion $\mcal{T}$, a (connected) subregion $\mcal{A}$, such that $\mcal{T} \subseteq \mcal{A} \subseteq \mcal{V}$, is selected. 
The two subregions $\mcal{T}$ and $\mcal{A}$ are called the ``target region'' and ``sum region,'' respectively.
For the sum region, a conditional distribution on $P(\bm{x} \mid \beta)$ is considered as
\begin{align}
P(\bm{x}_{\mcal{A}} \mid \bm{x}_{\partial \mcal{A}}; \beta) = \frac{P(\bm{x}\mid \beta)}{\sum_{\bm{x}_{\mcal{A}}}P(\bm{x}\mid \beta)},
\label{eqn:def_conditional_distribution_SMCI}
\end{align} 
where $\partial \mcal{A}$ (called the ``sample region'') denotes the first-nearest-neighboring region of $\mcal{A}$, defined by  
$\partial \mcal{A}:=\{i \mid (i,j) \in \mcal{E},\> j \in \mcal{A},\> i \not\in \mcal{A}\}$.
This conditional distribution can be immediately obtained as follows. 
The energy function in equation (\ref{eqn:Hamiltonian}) can be decomposed into two parts as 
\begin{align}
E(\bm{x}) = E_{\mcal{A}}(\bm{x}_{\mcal{A}}, \bm{x}_{\partial \mcal{A}}) 
+  E_{\mcal{A}^*}(\bm{x}_{\mcal{A}^*}),
\label{eqn:Hamiltonian_decompose}
\end{align}  
where $E_{\mcal{A}}(\bm{x}_{\mcal{A}}, \bm{x}_{\partial \mcal{A}})$ is the energy including all terms related to $\bm{x}_{\mcal{A}}$ 
and $E_{\mcal{A}^*}(\bm{x}_{\mcal{A}^*})$ is the energy unrelated to $\bm{x}_{\mcal{A}}$; 
here, $\mcal{A}^*$ is the complementary set of $\mcal{A}$. 
Using the decomposition of equation (\ref{eqn:Hamiltonian_decompose}), the conditional distribution in equation (\ref{eqn:def_conditional_distribution_SMCI}) is obtained as
\begin{align}
P(\bm{x}_{\mcal{A}} \mid \bm{x}_{\partial \mcal{A}}; \beta) \propto \exp\big( -\beta E_{\mcal{A}}(\bm{x}_{\mcal{A}}, \bm{x}_{\partial \mcal{A}}) \big).
\label{eqn:conditional_distribution_SMCI}
\end{align}

In SMCI, with the sample set $\mathbb{S}$ generated from $P(\bm{x} \mid \beta)$, the expectation is approximated by 
\begin{align}
\ave{f(\bm{x}_{\mcal{T}})}_{\beta} \approx
 \frac{1}{N}\sum_{\mu=1}^N\sum_{\bm{x}_{\mcal{A}}}f(\bm{x}_{\mcal{T}}) P\big(\bm{x}_{\mcal{A}} \mid \mbf{s}_{\partial \mcal{A}}^{(\mu)} ; \beta\big),
\label{eqn:GSMCI}
\end{align}
where $\mbf{s}_{\partial \mcal{A}}^{(\mu)}$ is the $\mu$th sampling point corresponding to the sample region. 
The relationship between the subregions is illustrated in figure \ref{fig:region_sheme_SMCI}. 
Two important properties of SMCI have been proved~\cite{SMCI2015,SMCI2020}: for a given $\mathbb{S}$, (i) SMCI is statistically more accurate than the standard MCI of equation (\ref{eqn:expectation-f_MCI})
and (ii) the approximation accuracy of SMCI monotonically increases as the size of the selected sum region increases. 
The simplest version of SMCI is the first-order SMCI (1-SMCI) method~\cite{SMCI2015}, in which the sum region is identical to the target region. 
The above two properties are maintained in general Markov random fields, including higher-order cases~\cite{SMCI2015,SMCI2020}. 
An example of the 1-SMCI method is presented in Appendix \ref{app:1-SMCI}.

However, SMCI has certain fundamental drawbacks.
It requires the execution of multiple summations (or integrations) over the sum region. 
Therefore, the sum region cannot easily expand in dense graphs; 
only the 1-SMCI and semi-second-order SMCI~\cite{SMCI2020} methods are applicable in dense graphs. 
The 1-SMCI method cannot be used when the target region is significantly large, 
with the exception of some special cases (e.g., when the target region is a tree).

The performances of MCI and SMCI are strongly dependent on the sampling quality. 
They degrade when a given sample set includes an unexpected bias. 
Therefore, the approximations in equations (\ref{eqn:expectation-f_MCI}) and (\ref{eqn:GSMCI}) 
would be poor in cases where it is difficult to perform high-quality sampling (i.e., a low-temperature case). 
In contrast, AIS, described in the following section, can reduce this type of performance degradation.

\subsection{Annealed importance sampling}
\label{sec:AIS}

AIS is a type of importance sampling based on MCMC with simulated annealing. 
In AIS, a sample set is generated as follows. 
First, for a sequence of the annealing schedule, $0=\beta_0 < \beta_1 < \cdots < \beta_K = 1$, 
set a sequence of distributions as
\begin{align}
P_k(\bm{x}) \propto P_0(\bm{x})^{1 - \beta_k}P(\bm{x} \mid \beta)^{\beta_k},
\label{eqn:AIS_Sequence_Pk}
\end{align}
where $P_0(\bm{x})$ is an initial (tractable) distribution, which is often set to a uniform distribution.
When $k = K$, distribution $P_k(\bm{x})$ is identified as $P(\bm{x} \mid \beta)$. 
Next, for $P_k(\bm{x})$, a transition probability $T_k(\bm{x}' \mid \bm{x})$, 
which satisfies the balance condition 
\begin{align}
P_k(\bm{x}') = \sum_{\bm{x}}T_k(\bm{x}' \mid \bm{x})P_k(\bm{x}) 
\label{eqn:DBC}
\end{align}
is defined. 
With the transition probability, generate the sequence of sample points $\mbf{X} = \{\mbf{x}^{(k)} \in \{-1,+1\}^n \mid k = 1,2,\ldots, K\}$ as  
\begin{align}
\begin{split}
\mbf{x}^{(1)}&\leftarrow P_0(\bm{x}),\\
\mbf{x}^{(k)}&\leftarrow T_{k-1}(\bm{x} \mid \mbf{x}^{(k-1)})\>\> (k = 2,3,\ldots,K).
\end{split}
\label{eqn:AIS_sampling}
\end{align}
The final point is employed as the sampled point, $\hat{\mbf{s}} =  \mbf{x}^{(K)}$, and the corresponding (unnormalized) importance weight is obtained by 
\begin{align}
\omega(\mbf{X}) := \prod_{k=1}^K \frac{P_k^{\dagger}(\mbf{x}^{(k)})}{P_{k-1}^{\dagger}(\mbf{x}^{(k)})}, 
\label{eqn:AIS_weight}
\end{align}
where $P_k^{\dagger}(\bm{x})$ is the relative probability of $P_k(\bm{x})$; i.e., $P_k(\bm{x}) = P_k^{\dagger}(\bm{x}) / Z_k$, 
where $Z_k$ is the partition function of $P_k(\bm{x})$.  
When the initial distribution is a uniform distribution, equation (\ref{eqn:AIS_weight}) is reduced to
\begin{align}
\omega(\mbf{X}) = \exp\Big(- \beta\sum_{k=1}^K (\beta_k - \beta_{k-1})E(\mbf{x}^{(k)})\Big).
\end{align}
By repeating the above procedure $N$ times, the sample set, $\mathbb{S}_{\mrm{AIS}} := \{ \hat{\mbf{s}}_{\mu}  \in \{-1,+1\}^n \mid \mu = 1,2,\ldots,N\}$, 
and the corresponding importance weights, $\{\omega_{\mu} \mid \mu = 1,2,\ldots,N\}$, are obtained.
With $\mathbb{S}_{\mrm{AIS}}$ and the importance weights, $\ave{f(\bm{x})}_{\beta}$ is approximated by
\begin{align}
\ave{f(\bm{x})}_{\beta} \approx \frac{1}{\Omega}\sum_{\mu=1}^N\omega_{\mu} f(\hat{\mbf{s}}_{\mu}), 
\label{eqn:expectation-f_AIS}
\end{align}
where $\Omega := \sum_{\mu=1}^N\omega_{\mu}$ is the partition function of AIS. 
A more detailed background of AIS is described in Appendix \ref{app:AIS_details}.

AIS can also approximate the free energy: $F(\beta) := - \beta^{-1}\ln Z(\beta)$~\cite{AIS2001,Salakhutdinov2008}, as 
\begin{align} 
F(\beta) \approx - \frac{1}{\beta} \ln Z_0 - \frac{1}{\beta} \ln \Big(\frac{\Omega}{N}\Big),
\label{eqn:FreeEnergy_AIS}
\end{align}
where $Z_0$ is the partition function of $P_0(\bm{x})$; therefore, $Z_0 = 2^n$ when $P_0(\bm{x})$ is a uniform distribution. 
This free-energy approximation is essentially the same as the method proposed by Jarzynski~\cite{Jarzynski1997}. 
The free-energy approximation based on AIS (or its variants) has also been actively developed in the field of machine learning~\cite{HAIS2012,RAISE2015,DS2015}. 
For the derivation of equation (\ref{eqn:FreeEnergy_AIS}), see equation (\ref{eqn:approximation_r(beta)}).

\subsection{Numerical experiment: AIS versus SMCI}
\label{sec:experiment_AIS_vs_SMCI}

Consider an Ising model with $n = 20$. On the Ising model, the approximation accuracies of AIS and the 1-SMCI method were investigated through numerical experiments. 
The accuracy was measured by the mean absolute error (MAE) of the covariances, $\chi_{i,j} = \ave{x_ix_j}_{\beta} - \ave{x_i}_{\beta}\ave{x_j}_{\beta}$, defined by  
\begin{align}
\frac{1}{|\mcal{E}|}\sum_{(i,j) \in \mcal{E}} \big| \chi_{i,j}^{\mrm{exact}} - \chi_{i,j}^{\mrm{approx}}\big|,
\label{eqn:MAE}
\end{align}
where $\chi_{i,j}^{\mrm{exact}}$ is the exact covariance and $\chi_{i,j}^{\mrm{approx}}$ is its approximation obtained from an approximation method. 
In AIS, the sequence of the annealing schedule was set as $\beta_k = k / K$ with $K = 1000$; 
furthermore, 1-step (asynchronous) Gibbs sampling was considered as the transition probability. 
The initial distribution of AIS was set to a uniform distribution.
Sample set $\mathbb{S}$ used in the 1-SMCI method was obtained using $N$ parallel Gibbs sampling with simulated annealing, 
whose annealing schedule was almost identical to that of AIS, i.e., 
a sampling point in $\mathbb{S}$ was generated using ancestral sampling: 
\begin{align*}
\mbf{x}^{(0)}\leftarrow P_0(\bm{x}),\quad 
\mbf{x}^{(k)}\leftarrow T_{k}(\bm{x} \mid \mbf{x}^{(k-1)})\>\> (k = 1,2,\ldots,K),
\end{align*}
and $\mbf{x}^{(K)}$ was then employed as the sampled point.
Therefore, the sampling costs of $\mathbb{S}_{\mrm{AIS}}$ and $\mathbb{S}$ were almost the same; 
additionally, $N = 1000$ was used for both $\mathbb{S}_{\mrm{AIS}}$ and $\mathbb{S}$.

\begin{figure*}[tb]
\centering
\includegraphics[height=5cm]{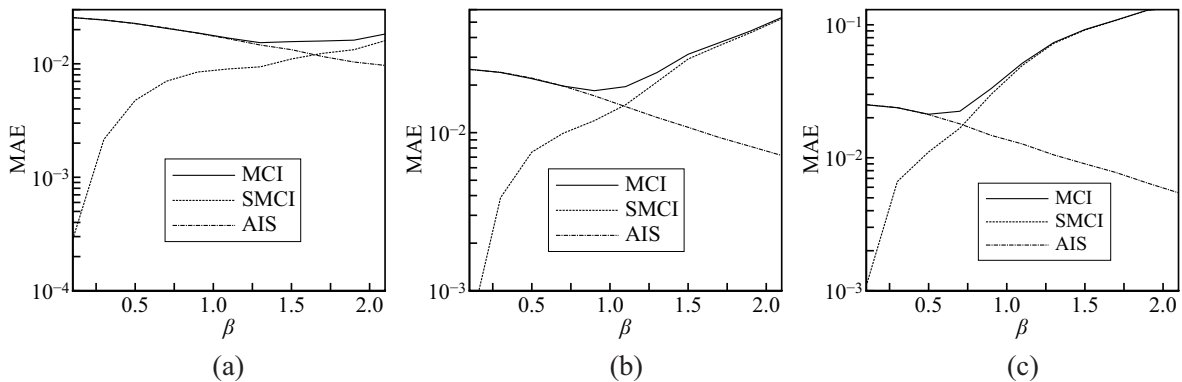}
\caption{MAE in equation (\ref{eqn:MAE}) versus the inverse temperature $\beta$ when (a) $p = 0.2$, (b) $p = 0.4$, and (c) $p = 0.8$. 
These plots present the averages over 1000 experiments.}
\label{fig:AIS_vs_SMCI}
\end{figure*}

Figure \ref{fig:AIS_vs_SMCI} depicts the results against the inverse temperature $\beta$ in the Ising model defined on a random graph with connection probability $p$. 
In the Ising model, $\{h_i\}$ and $\{J_{i,j}\}$ were randomly selected according to a uniform distribution over $[-1,+1]$.
For comparison, the results obtained using the standard MCI with $\mathbb{S}$ were also plotted. 
In the high-temperature region (i.e., the low $\beta$ region), the 1-SMCI method was significantly superior than the other methods. 
However, the accuracies of the 1-SMCI method and standard MCI were poor in the low-temperature region (i.e., the high $\beta$ region). 
This is because, in the low-temperature region, the quality of sampling tends to degrade; 
therefore, the obtained size-limited sample set cannot incorporate the detailed structure of the distribution.    
Meanwhile, it is noteworthy that AIS did not exhibit such degradation.  

\section{Proposed Method: AIS-based SMCI}
\label{sec:AIS+SMCI}

In this section, the proposed approximation method that combines AIS and SMCI is discussed. 
The experimental results from section \ref{sec:experiment_AIS_vs_SMCI} elucidated that SMCI is effective in high-temperature regions 
and AIS is effective in low-temperature regions. 
Combining both methods may provide a method that is effective over a broad range of temperature. 

Consider a function
\begin{align}
f_{(\mcal{T}:\mcal{A})}(\bm{x}_{\partial \mcal{A}}):=\sum_{\bm{x}_{\mcal{A}}}f(\bm{x}_{\mcal{T}}) P\big(\bm{x}_{\mcal{A}} \mid \bm{x}_{\partial \mcal{A}} ; \beta\big),
\end{align} 
whose conditional distribution can be expressed via equation (\ref{eqn:conditional_distribution_SMCI}). 
The expectation of this function is equivalent to $\ave{f(\bm{x}_{\mcal{T}})}_{\beta}$ because
\begin{align*}
\ave{f_{(\mcal{T}:\mcal{A})}(\bm{x}_{\partial \mcal{A}})}_{\beta} &= \sum_{\bm{x}} f_{(\mcal{T}:\mcal{A})}(\bm{x}_{\partial \mcal{A}})P(\bm{x} \mid \beta)\nn
&=\sum_{\bm{x}} f(\bm{x}_{\mcal{T}})P(\bm{x} \mid \beta).
\end{align*}
Equation (\ref{eqn:GSMCI}) can be considered as the approximation of $\ave{f_{(\mcal{T}:\mcal{A})}(\bm{x}_{\partial \mcal{A}})}_{\beta}$ 
based on the standard MCI of equation (\ref{eqn:expectation-f_MCI}). 
Based on the AIS of equation (\ref{eqn:expectation-f_AIS}), instead of the standard MCI, the following approximation can be obtained:
\begin{align}
\ave{f(\bm{x}_{\mcal{T}})}_{\beta} &\approx \frac{1}{\Omega}\sum_{\mu=1}^N\omega_{\mu} f_{(\mcal{T}:\mcal{A})}(\hat{\mbf{s}}_{\partial \mcal{A}}^{(\mu)})\nn
&= \frac{1}{\Omega} 
\sum_{\mu=1}^N\omega_{\mu}\sum_{\bm{x}_{\mcal{A}}}f(\bm{x}_{\mcal{T}}) P\big(\bm{x}_{\mcal{A}} \mid \hat{\mbf{s}}_{\partial \mcal{A}}^{(\mu)} ; \beta\big),
\label{eqn:expectation-f_SMCI+AIS}
\end{align}
where $\mathbb{S}_{\mrm{AIS}} = \{ \hat{\mbf{s}}_{\mu}  \in \{-1,+1\}^n \mid \mu = 1,2,\ldots,N\}$ 
and $\{\omega_{\mu} \mid \mu = 1,2,\ldots,N\}$ represents the sample set of AIS and the corresponding importance weights, respectively, 
which have been explained in Section \ref{sec:AIS}; 
$\Omega$ is the partition function of AIS and $\hat{\mbf{s}}_{\partial \mcal{A}}^{(\mu)}$ is the $\mu$th sampling point corresponding to the sample region of SMCI. 
Equation (\ref{eqn:expectation-f_SMCI+AIS}) denotes the method proposed in this study. 

In the following, the efficiency of the proposed method is considered.
As described in equation (\ref{eqn:Variance_AIS}), the asymptotic variance of the approximation of $\ave{f(\bm{x}_{\mcal{T}})}_{\beta}$ 
using AIS is approximated as~\cite{AIS2001} 
\begin{align}
V_{\mrm{AIS}}[f(\bm{x}_{\mcal{T}})] \approx \frac{1}{N} W V_{\beta}[f(\bm{x}_{\mcal{T}})],
\label{eqn:Variance_AIS_TargetRegion}
\end{align}
where $V_{\beta}[f(\bm{x}_{\mcal{T}})]:= \ave{f(\bm{x}_{\mcal{T}})^2}_{\beta} -\ave{f(\bm{x}_{\mcal{T}})}_{\beta}^2$ is the variance of $f(\bm{x}_{\mcal{T}})$ 
and $W\geq 1$ is the constant factor that is independent of $f(\bm{x}_{\mcal{T}})$.
This asymptotic variance indicates the efficiency of this approximation (evidently, a lower variance is better).
The factor $W$ may be expected to be close to $1$ when $P(\bm{x} \mid \beta)$ has few isolated modes (namely, when $\beta$ is not large). 
When a given sample set, $\mathbb{S}$, does not include an unexpected bias, 
the asymptotic variance of the standard MCI for $\ave{f(\bm{x}_{\mcal{T}})}_{\beta}$ is expressed as $V_{\mrm{MCI}}[f(\bm{x}_{\mcal{T}})]:=N^{-1}V_{\beta}[f(\bm{x}_{\mcal{T}})]$. 
Therefore, in cases where high-quality sampling can be executed, the efficiency of AIS is considered to be almost the same as that of the standard MCI; 
in fact, the accuracies of both methods were almost the same in the high-temperature region in the numerical results presented in section \ref{sec:experiment_AIS_vs_SMCI}. 
In contrast, in the low-temperature region, 
the accuracy of MCI significantly degraded owing to the degradation of the sampling quality, whereas that of AIS did not.

This argument can be extended to the proposed method in equation (\ref{eqn:expectation-f_SMCI+AIS}). 
The asymptotic variance of the proposed method can be estimated as
\begin{align}
V_{\mrm{SMCI+AIS}}[f(\bm{x}_{\mcal{T}})] \approx \frac{1}{N} W V_{\beta}[f_{(\mcal{T}:\mcal{A})}(\bm{x}_{\partial \mcal{A}})].
\label{eqn:Variance_SMCI+AIS_TargetRegion}
\end{align}
The asymptotic variance of SMCI is $V_{\mrm{SMCI}}[f(\bm{x}_{\mcal{T}})]:=N^{-1}V_{\beta}[f_{(\mcal{T}:\mcal{A})}(\bm{x}_{\partial \mcal{A}})]$, 
which was proved to be $V_{\mrm{SMCI}}[f(\bm{x}_{\mcal{T}})]\leq V_{\mrm{MCI}}[f(\bm{x}_{\mcal{T}})]$~\cite{SMCI2015,SMCI2020}. 
Using equations (\ref{eqn:Variance_AIS_TargetRegion}) and (\ref{eqn:Variance_SMCI+AIS_TargetRegion}) and this inequality, 
\begin{align}
V_{\mrm{SMCI+AIS}}[f(\bm{x}_{\mcal{T}})] \leq V_{\mrm{AIS}}[f(\bm{x}_{\mcal{T}})] 
\label{eqn:Variance_Bound}
\end{align}
is obtained, which implies that the proposed method is more efficient than the standard AIS. 

\begin{figure}[tb]
\centering
\includegraphics[height=4cm]{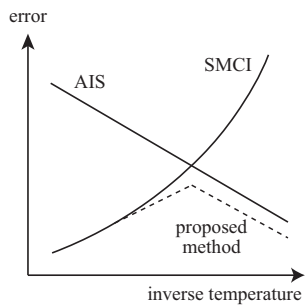}
\caption{Qualitative illustration of the expected performance of the proposed method.}
\label{fig:accuracy_image}
\end{figure}

Based on the above arguments, the following two properties can be expected: 
the accuracy of the proposed method is (i) almost the same as that of SMCI in high-temperature regions 
and (ii) higher than that of AIS in low-temperature regions.
If these properties are satisfied, a result similar to that illustrated in figure \ref{fig:accuracy_image} can be obtained. 
The empirical justification of this expectation is demonstrated in the following section. 

The proposed method and AIS require $O(KN)$ steps of Gibbs sampling to generate the set of sampling points, $\{\hat{\mbf{s}}_{\mu}  \mid \mu = 1,2,\ldots,N\}$, 
and that of the corresponding importance weights, $\{\omega_{\mu} \mid \mu = 1,2,\ldots,N\}$, 
when 1-step Gibbs sampling is employed as the transition probability, $T_k(\bm{x}' \mid \bm{x})$. 
Because $N$ different sequences of Gibbs sampling can be performed independently, the implementation of these sequences can be easily parallelized.

\section{Numerical Experiment}
\label{sec:experiment_AIS+SMCI}

In this section, the performance of the proposed method is examined using numerical experiments. 
In the following experiments, the term ``SMCI'' denotes the 1-SMCI method. 
For the detailed formulation of the 1-SMCI method, see Appendix \ref{app:1-SMCI}.

\subsection{Ising model on random graph}
\label{sec:experiment_random}

\begin{figure*}[tb]
\centering
\includegraphics[height=5cm]{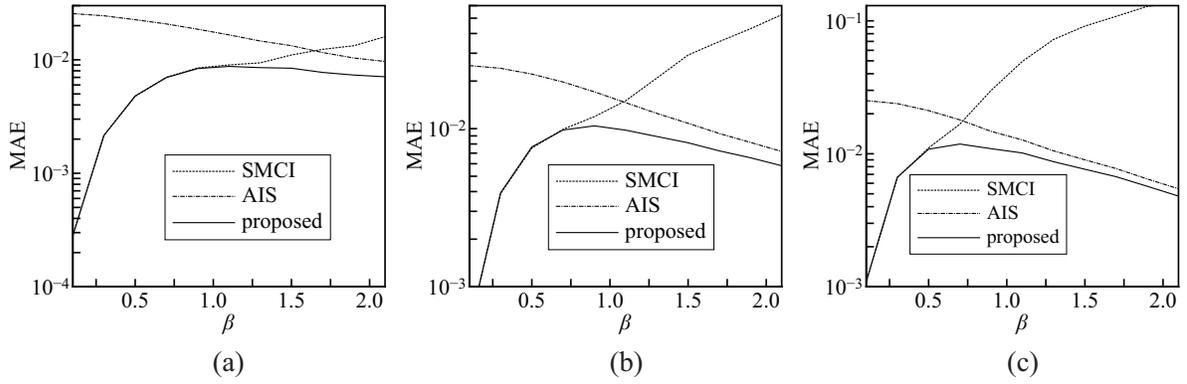}
\caption{MAE in equation (\ref{eqn:MAE}) versus $\beta$ when (a) $p = 0.2$, (b) $p = 0.4$, and (c) $p = 0.8$. 
The results of SMCI and AIS are identical to those in figure \ref{fig:AIS_vs_SMCI}.
These plots present the averages over 1000 experiments.}
\label{fig:PM_AIS_vs_SMCI}
\end{figure*}

The validation of the proposed method is demonstrated via numerical experiments,  
whose settings are the same as those in the numerical experiments presented in section \ref{sec:experiment_AIS_vs_SMCI}, unless otherwise noted.  
Figure \ref{fig:PM_AIS_vs_SMCI} depicts the results obtained from the proposed method, 
in which the setting of the experiment is identical to that of figure \ref{fig:AIS_vs_SMCI}. 
The accuracy of the proposed method was consistent with the expected results illustrated in figure \ref{fig:accuracy_image}. 
The proposed method is efficient in both high- and low-temperature regions. 

\begin{figure*}[tb]
\centering
\includegraphics[height=10cm]{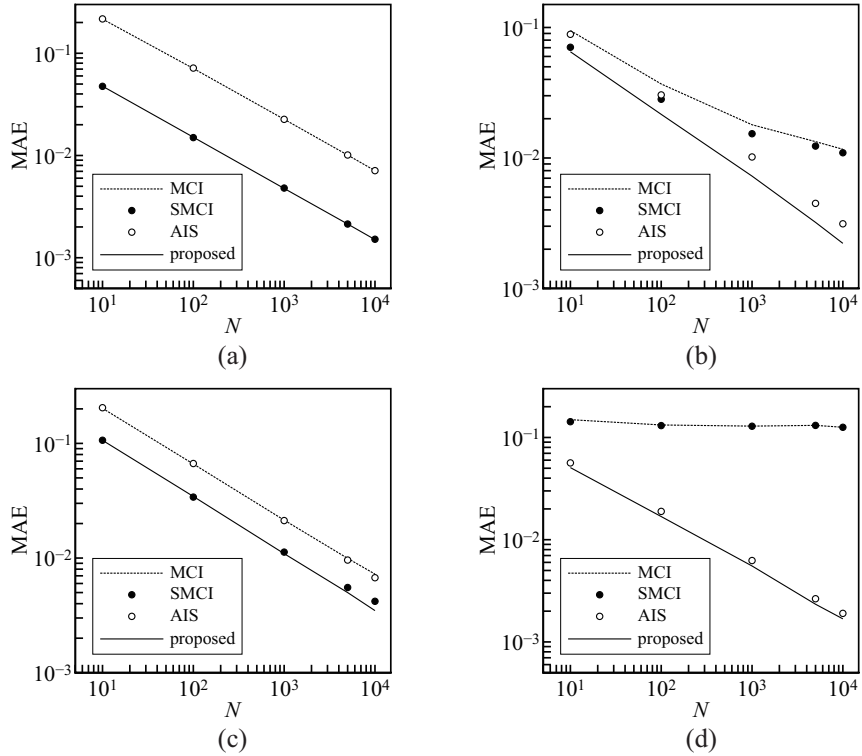}
\caption{MAE in equation (\ref{eqn:MAE}) versus $N$ when (a) $p = 0.2$ and $\beta = 0.5$, (b) $p = 0.2$ and $\beta = 2$, 
(c) $p = 0.8$ and $\beta = 0.5$, and (d) $p = 0.8$ and $\beta = 2$.
These plots present the averages over 1000 experiments.}
\label{fig:N_dependency}
\end{figure*}

\begin{figure*}[tb]
\centering
\includegraphics[height=10cm]{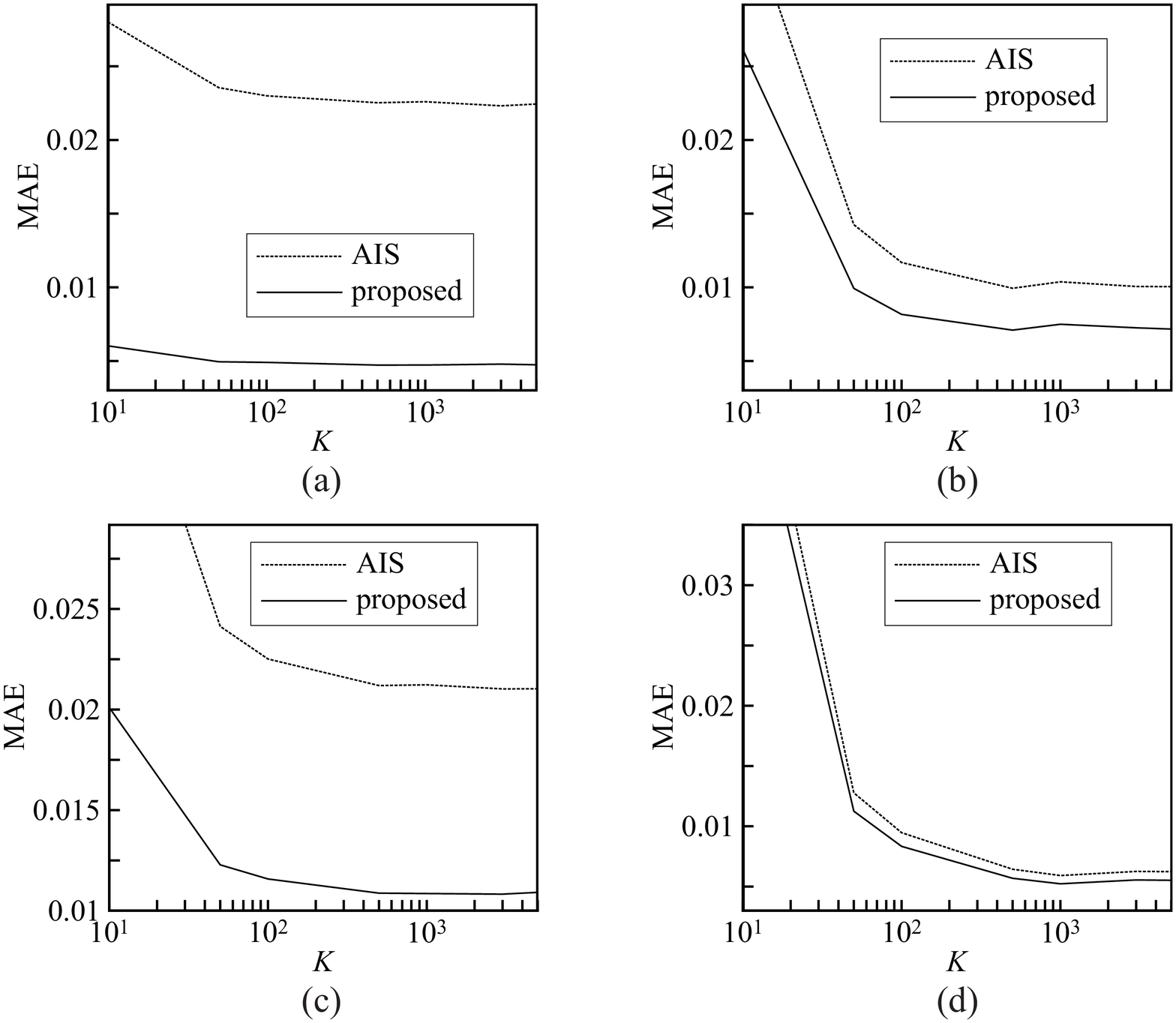}
\caption{MAE in equation (\ref{eqn:MAE}) versus $K$ when (a) $p = 0.2$ and $\beta = 0.5$, (b) $p = 0.2$ and $\beta = 2$, 
(c) $p = 0.8$ and $\beta = 0.5$, and (d) $p = 0.8$ and $\beta = 2$.
These plots present the averages over 1000 experiments.}
\label{fig:K_dependency}
\end{figure*}

In the following, the dependency of the proposed method on $N$ and $K$, the sizes of the sample set and annealing sequence, respectively, are investigated. 
Figure \ref{fig:N_dependency} depicts the results against $N$, in which $K = 1000$ was fixed. 
The errors of AIS and the proposed method decreased at a speed approximately proportional to $O(N^{-1/2})$ in both high- and low-temperature cases; 
however, those of MCI and SMCI did not exhibit such a decrease in the low-temperature cases (figures \ref{fig:N_dependency}(b) and (d)), 
which can be attributed to the unexpected bias in $\mathbb{S}$. 
Figure \ref{fig:K_dependency} depicts the results against $K$, in which $N = 1000$ was fixed. 
The errors decreased as $K$ increased; they became saturated at approximately $K = 500$; 
thus, $K = 1000$ seems to be sufficient in the presented experiments.

\subsection{Hopfield-type and bipartite Ising models}
\label{sec:experiment_HFM&RBM}

In this section, the results of numerical experiments on a Hopfield-type Ising model~\cite{HFM1982} 
and bipartite Ising model are presented. In these experiments $K = 1000$ and $N = 1000$ were used.

\begin{figure*}[tb]
\centering
\includegraphics[height=5cm]{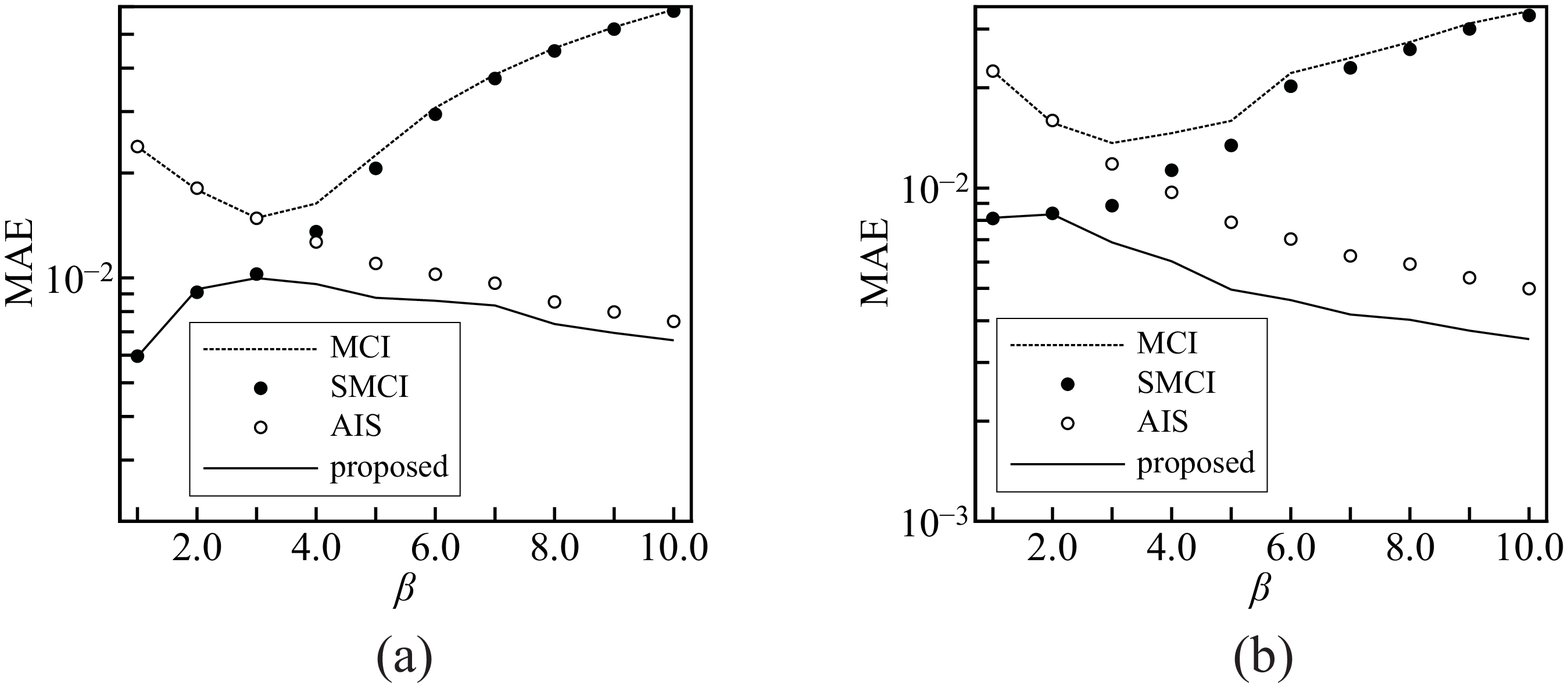}
\caption{Results on the Hopfield-type Ising models; MAE in equation (\ref{eqn:MAE}) versus $\beta$ when (a) $\alpha = 0.2$ and (b) $\alpha = 0.5$. 
These plots present the averages over 2000 experiments.}
\label{fig:HFM}
\end{figure*}

First, a Hopfield-type Ising model~\cite{HFM1982} was considered, 
in which the interactions $J_{i,j}$ were determined by
\begin{align*}
J_{i,j} = \frac{1}{n}\sum_{k = 1}^m \xi_{i,k}\xi_{j,k}, 
\end{align*}
where $\bm{\xi} = \{\xi_{i,k}\in \{-1,+1\} \mid i \in \mcal{V},\, k = 1,2,\ldots,m\}$ were randomly generated, 
and the biases $h_i$ were set to zero. 
Figure \ref{fig:HFM} depicts the results on the Hopfield-type Ising models with $n = 20$, for $\alpha := m / n = 0.2, \, 0.5$.

\begin{figure*}[tb]
\centering
\includegraphics[height=5cm]{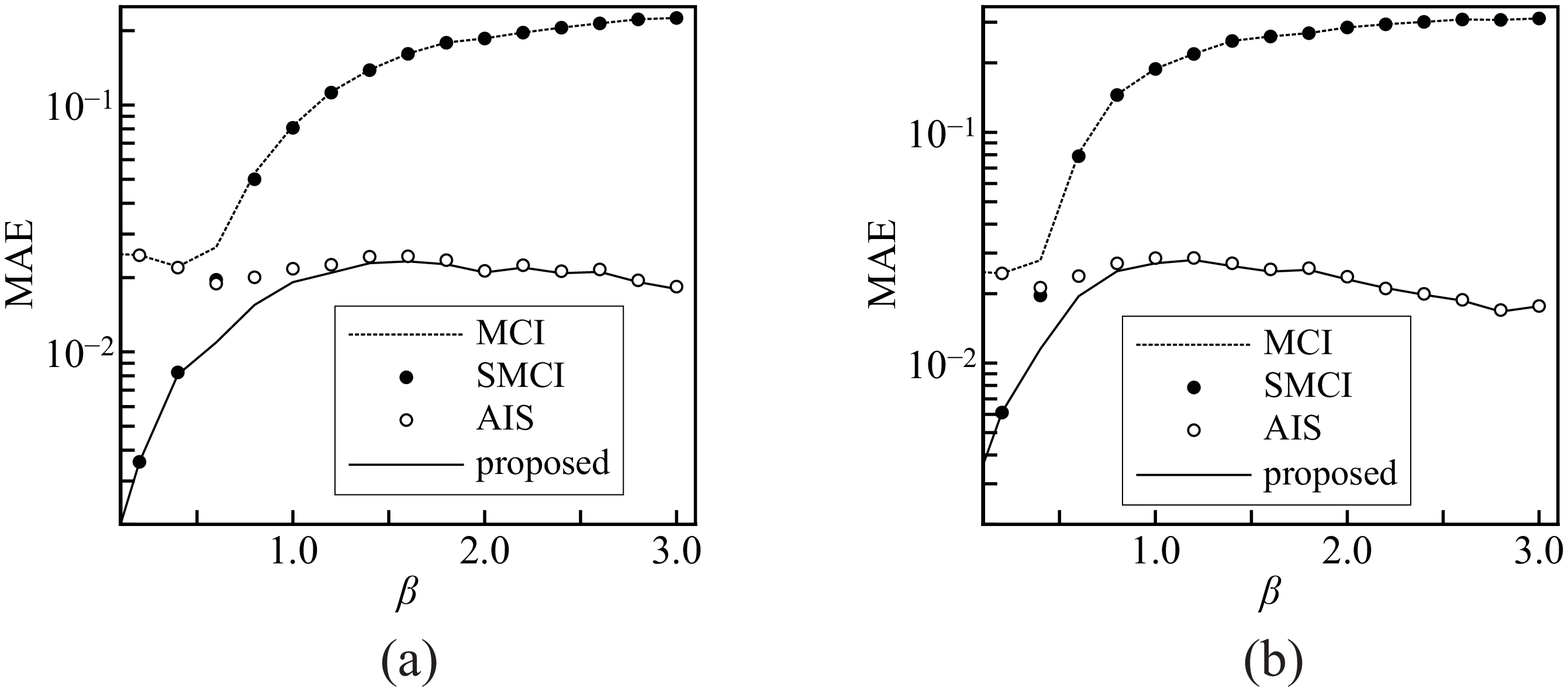}
\caption{Results on the bipartite Ising models; MAE in equation (\ref{eqn:MAE}) versus $\beta$ when (a) $p = 0.5$ and (b) $p = 1$. 
These plots present the averages over 3000 experiments.}
\label{fig:RBM}
\end{figure*}

Next, an Ising model defined on a bipartite graph was considered. 
This model is related to restricted Boltzmann machines in the field of machine learning~\cite{CD2002,SRBM2017}. 
The set of vertices $\mcal{V}$ is divided into two different groups (or layers) $\mcal{V}_0$ and $\mcal{V}_1$; 
a variable in one group interacts with variables in the other group with probability $p$ 
and does not interact with variables in the same group.  
Figure \ref{fig:RBM} depicts the results on the bipartite Ising models with $|\mcal{V}_0| = 10$ and $|\mcal{V}_1| = 100$ (i.e., $n = 110$), 
in which the biases and interactions were generated in the same manner as that in section \ref{sec:experiment_AIS_vs_SMCI}. 
The transition probability was based on a group-wise blocked Gibbs sampling, i.e., 
\begin{align*}
T_k(\bm{x}' \mid \bm{x})= P_k(\bm{x}'_{\mcal{V}_1} \mid \bm{x}'_{\mcal{V}_0})P_k(\bm{x}'_{\mcal{V}_0} \mid \bm{x}_{\mcal{V}_1}).
\end{align*}
It is noteworthy that exact expectations on the model can be evaluated in $O(2^{|\mcal{V}_0|})$-time 
through the marginalization: 
\begin{align*}
&P(\bm{x}_{\mcal{V}_0} \mid \beta) = \sum_{\bm{x}_{\mcal{V}_1}}P(\bm{x} \mid \beta) \nn
&\propto
\exp\Big\{ \sum_{i \in \mcal{V}_0} b_i x_i + \sum_{j \in \mcal{V}_1} \ln \cosh\Big(b_j + \sum_{k \in \mcal{V}_0} J_{k,j}x_k\Big)\Big\},
\end{align*}
where $J_{k,j} = 0$ if $x_k\, (k \in \mcal{V}_0)$ and $x_j\, (k \in \mcal{V}_1)$ have no interactions. 

The proposed method preformed most efficiently in both models.

\subsection{Comparison with parallel tempering} 

\begin{figure*}[tb]
\centering
\includegraphics[height=5cm]{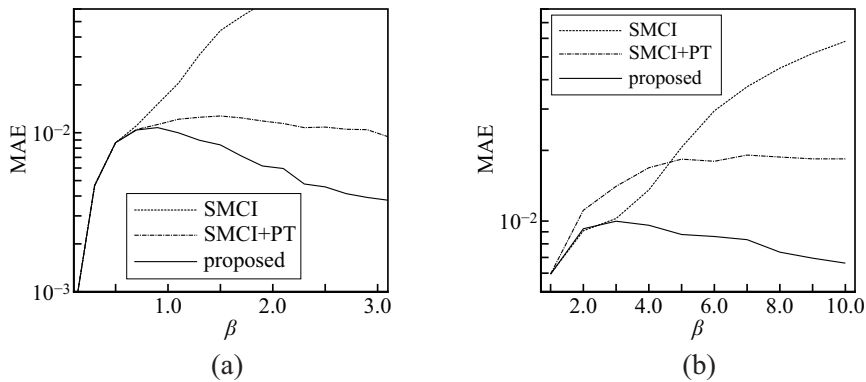}
\caption{Comparison with PT-based method; MAE in equation (\ref{eqn:MAE}) versus $\beta$ on (a) the random graph with $p = 0.5$ 
and (b) the Hopfield-type Ising models with $\alpha = 0.2$. 
The results of SMCI and the proposed method in panel (b) are identical to those in figure \ref{fig:HFM}(a).
These plots present the averages over 2000 experiments.}
\label{fig:vs_PT}
\end{figure*}

This section describes the comparison of the proposed method (i.e., AIS-based SMCI) with PT (i.e., PT-based SMCI). 
In the PT-based method, ten different temperature processes, $1 = \beta_1 > \beta_2 > \cdots > \beta_{10} = 0.01$ and 
the sampling interval of 100 MC steps were used to coincide with the proposed method in terms of the number of MC steps; 
the temperature intervals were set according to a geometric sequence.
Figure \ref{fig:vs_PT} depicts the results on (a) the random graph with $p = 0.5$ and (b) the Hopfield-type Ising models with $\alpha = 0.2$, 
respectively. 
In these experiments, $n = 20$, $K = 1000$ and $N = 1000$ were used. 
The setting of parameters (biases and interactions) were the same as those in sections \ref{sec:experiment_AIS_vs_SMCI} and \ref{sec:experiment_HFM&RBM}, 
respectively. 
The PT-based method (``SMCI+PT'' in figure \ref{fig:vs_PT}) improves the accuracy in the low-temperature region. 
However, the proposed method is more efficient.

\subsection{Computational efficiency}

In this section, the comparison of the proposed method (based on the 1-SMCI method) with AIS in terms of the computational cost for 
evaluating $\ave{x_i}_{\beta}$ for all $i \in \mcal{V}$ and $\ave{x_ix_j}_{\beta}$ for all $(i,j) \in \mcal{E}$ is discussed. 
Assume that $|\mcal{E}| > |\mcal{V}| \gg 1$, $N\gg 1$, and $K\gg 1$. The cost of one-step Gibbs sampling can be estimated as $O(|\mcal{E}|)$; 
therefore, the cost for generating $\mathbb{S}_{\mrm{AIS}}$ is $O(KN|\mcal{E}|)$. 
The cost for evaluating the importance weights $\{\omega_{\mu}\}$ is also $O(KN|\mcal{E}|)$; 
here, the cost of $O(|\mcal{E}|)$ is the evaluation cost of the energy function of equation (\ref{eqn:Hamiltonian}). 
Given $\mathbb{S}_{\mrm{AIS}}$ and $\{\omega_{\mu}\}$, 
the cost for evaluating equation (\ref{eqn:expectation-f_AIS}) for all expectations (i.e., $\{\ave{x_i}_{\beta}\}$ and $\{\ave{x_ix_j}_{\beta}\}$)
can be estimated as $O(N |\mcal{E}|)$. 
The above arguments conclude that the total cost of AIS is $O(KN|\mcal{E}|)$. 

The total cost of the proposed method is the same as that of AIS in the perspective of the order; 
because, given $\mathbb{S}_{\mrm{AIS}}$ and $\{\omega_{\mu}\}$, the cost for evaluating equation (\ref{eqn:expectation-f_SMCI+AIS}) for all expectations 
can be estimated as $O(N |\mcal{E}|)$ (see Appendix \ref{app:1-SMCI}), which is the same as that of AIS. 

The costs for evaluating all expectations of both methods are the same as mentioned above. 
However, the evaluation of the proposed method is more time consuming than that of AIS in terms of the computational time (or CPU time), 
because it requires more complicated computations. 
Table \ref{tab:time} presents the computational times of AIS and the proposed method in an Ising model on a complete graph 
obtained from our implementation without a parallel computation, in which $K = 1000$ and $N = 1000$. 
The computational time for evaluating the expectation of the proposed method is tens of times slower than that of AIS. 
However, these computational times are considerably small compared with those required for sampling and evaluating the importance weights.

\begin{table*}[htb]
\caption{Comparison of the computational time; ``sampling'' denotes the time required for generating $\mathbb{S}_{\mrm{AIS}}$; 
``wights'' denotes the time required for evaluating the importance weights $\{\omega_{\mu}\}$; and 
``expectations'' denotes the time required for evaluating all expectations. 
The values in this table are the average times over 10 experiments. }
\label{tab:time}
\begin{tabular}{|c|c|c|c|c|c|}
\hline
$n$                  & method   & sampling [ms]            & weights [ms]             & expectations [ms]   & total [s] \\ \hline
\multirow{2}{*}{50}  & AIS      & \multirow{2}{*}{14493.2} & \multirow{2}{*}{613.2}   & 2                   & 15.11       \\ \cline{2-2} \cline{5-6} 
                     & proposed &                          &                          & 88.4                & 15.19       \\ \hline
\multirow{2}{*}{100} & AIS      & \multirow{2}{*}{36127.2} & \multirow{2}{*}{2144}    & 10                  & 38.28       \\ \cline{2-2} \cline{5-6} 
                     & proposed &                          &                          & 371.2               & 38.64       \\ \hline
\multirow{2}{*}{200} & AIS      & \multirow{2}{*}{102600}  & \multirow{2}{*}{10483.7} & 68.5                & 113.15      \\ \cline{2-2} \cline{5-6} 
                     & proposed &                          &                          & 1541.4              & 114.63      \\ \hline
\end{tabular}
\end{table*}

\section{Summary and Future Studies}
\label{sec:summary}

In this study, a new effective sampling approximation, AIS-based SMCI, was proposed to evaluate the expectations on an Ising model. 
As demonstrated by the numerical results in section \ref{sec:experiment_AIS+SMCI}, 
the importance weights of AIS considerably improved the approximation performance of SMCI in the low-temperature region. 
Because the proposed method does not use any characteristic property of the Ising model (at least in theory), 
it can be applied to more general models besides the Ising model, such as a high-order Markov random field.  

The proposed method performed efficiently in both high- and low-temperature regions 
without using a sophisticated sampling method, besides Gibbs sampling; 
this is a significant result in terms of cost and implementation. 
However, the consideration of alternative possibilities is still required. 
SMCI does not have any limitation in terms of the sampling method; 
therefore, SMCI can be directly combined with more sophisticated sampling methods, 
such as the Suwa-Todo method~\cite{ST2010} and belief-propagation-guided MCMC~\cite{BP+MCMC2014}. 
This can be an interesting future investigation.
Furthermore, the improvement of AIS must also be considered.
Hukushima and Iba proposed a resampling method for AIS that can reduce the variance of the importance weights~\cite{resamplingAIS2003}; 
we believe that the resampling method can improve the performance of the proposed method.

As mentioned in introduction, accurate approximations of expectations on Ising models are also required in the field of machine learning. 
The application of the proposed method to the Boltzmann-machine learning and inference will be addressed in our future project. 

\appendix

\section{1-SMCI method}
\label{app:1-SMCI}

This appendix shows the formulations of the 1-SMCI method for $\ave{x_i}_{\beta}$ and $\ave{x_i x_j}_{\beta}$ 
for the sample set $\mathbb{S}$~\cite{SMCI2015,SMCI2020}.

For the approximation of $\ave{x_i}_{\beta}$, the target and sum regions are set as $\mcal{T} = \mcal{A} = \{i\}$; 
thus, the conditional distribution in equation (\ref{eqn:conditional_distribution_SMCI}) is 
\begin{align}
P(x_i \mid \bm{x}_{\partial \{i\}}; \beta) \propto \exp \beta \Big( h_i x_i + \sum_{j \in \partial \{i\}} J_{i,j} x_i x_j\Big),
\label{eqn:conditional_distribution_1SMCI_xi}
\end{align}
where $\partial \{i\}$ is the first-nearest-neighboring region of $i$. 
Equations (\ref{eqn:GSMCI}) and (\ref{eqn:conditional_distribution_1SMCI_xi}) lead to 
\begin{align}
\ave{x_i}_{\beta}&\approx \frac{1}{N}\sum_{\mu=1}^N \sum_{x_i }x_iP(x_i \mid \bm{x}_{\partial \{i\}}; \beta)\nn
&=\frac{1}{N}\sum_{\mu=1}^N \tanh \phi_{i}^{(\mu)},
\label{eqn:1-SMCI_xi}
\end{align}
where
\begin{align*} 
\phi_{i}^{(\mu)}:= \beta h_i + \beta\sum_{j \in \partial \{i\}} J_{i,j} \mrm{s}_j^{(\mu)};
\end{align*} 
here, $\mrm{s}_j^{(\mu)}$ is the $\mu$th sampling point corresponding to vertex $j$. 

For the approximation of $\ave{x_i x_j}_{\beta}$, the target and sum regions are set as $\mcal{T} = \mcal{A} = \{i, j\}$; 
thus, the conditional distribution in equation (\ref{eqn:conditional_distribution_SMCI}) is  
\begin{align}
&P(x_i,x_j \mid \bm{x}_{\partial \{i,j\}}; \beta) \propto \exp \beta \Big( h_i x_i + h_j x_j  +J_{i,j}x_i x_j \nn
\aleq
+ \sum_{k \in \partial \{i\} \setminus \{j\}} J_{i,k} x_i x_k + \sum_{l \in \partial \{j\} \setminus \{i\}} J_{j,l} x_j x_l\Big).
\label{eqn:conditional_distribution_1SMCI_xixj}
\end{align}
Equations (\ref{eqn:GSMCI}) and (\ref{eqn:conditional_distribution_1SMCI_xixj}) lead to 
\begin{align}
\ave{x_ix_j}_{\beta}&\approx \frac{1}{N}\sum_{\mu=1}^N\sum_{x_i,x_j}x_ix_jP(x_i,x_j \mid \bm{x}_{\partial \{i,j\}}; \beta)\nn
&= \frac{1}{N}\sum_{\mu=1}^N \tanh \big[ \atanh\big\{ \tanh\big(\psi_{i:j}^{(\mu)}\big)\tanh\big(\psi_{j:i}^{(\mu)}\big)\big\}\nn
\aleq
+ \beta J_{i,j}\big],
\label{eqn:1-SMCI_xixj}
\end{align}
where $\psi_{i:j}^{(\mu)} := \phi_{i}^{(\mu)} - \beta J_{i,j} \mrm{s}_j^{(\mu)}$ and $\atanh$ is the inverse hyperbolic tangent function.

For the given $\mathbb{S}$, the computational cost for evaluating $\{\phi_{i}^{(\mu)}\}$ is $O(N |\mcal{E}|)$; 
and for the given $\{\phi_{i}^{(\mu)}\}$, the costs for evaluating equation (\ref{eqn:1-SMCI_xi}) for a specific $i \in \mcal{V}$ 
and equation (\ref{eqn:1-SMCI_xixj}) for a specific $(i,j) \in \mcal{E}$ are $O(N)$.
Therefore, for the given $\mathbb{S}$, the total computational cost for evaluating equation (\ref{eqn:1-SMCI_xi}) for all $i \in \mcal{V}$ 
and equation (\ref{eqn:1-SMCI_xixj}) for all $(i,j) \in \mcal{E}$ can be estimated as $O(N |\mcal{E}|)$.

\section{Details of annealed importance sampling}
\label{app:AIS_details}

First, the background of AIS described in section \ref{sec:AIS} is considered.
The expectation $\ave{f(\bm{x})}_{\beta}$ is rewritten as 
\begin{align}
\ave{f(\bm{x})}_{\beta} = \sum_{\bm{X}} \omega_{\mrm{norm}}(\bm{X}) f(\bm{x}^{(K)}) Q_{\mrm{f}}(\bm{X}),
\label{eqn:expectation_f_extension}
\end{align}
where $\bm{X} = \{\bm{x}^{(k)} \in \{-1,+1\}^n\mid k = 1,2,\ldots, K\}$ and
\begin{align}
\omega_{\mrm{norm}}(\bm{X}):= \frac{Q_{\mrm{b}}(\bm{X})}{Q_{\mrm{f}}(\bm{X})}
\label{eqn:AIS_weight_normalized}
\end{align}
is the (normalized) importance weight. 
Here, the two distributions, $Q_{\mrm{f}}(\bm{X})$ and $Q_{\mrm{b}}(\bm{X})$, are defined as follows:
\begin{align}
Q_{\mrm{f}}(\bm{X})&:= P_0(\bm{x}^{(1)})\prod_{k=1}^{K-1} T_k(\bm{x}^{(k+1)} \mid \bm{x}^{(k)}), 
\label{eqn:forward_process}\\
Q_{\mrm{b}}(\bm{X})&:= P_K(\bm{x}^{(K)})\prod_{k=1}^{K-1} \tilde{T}_k(\bm{x}^{(k)} \mid \bm{x}^{(k+1)}),
\label{eqn:backward_process}
\end{align}
where $P_0(\bm{x})$ and $P_K(\bm{x}) = P(\bm{x} \mid \beta)$ are the initial and target distributions, respectively, 
and $T_k(\bm{x}' \mid \bm{x})$ is the transition probability. 
Here, $\tilde{T}_k(\bm{x} \mid \bm{x}')$ is the ``reverse'' transition probability, satisfying 
\begin{align*}
\tilde{T}_k(\bm{x} \mid \bm{x}')=\frac{T_k(\bm{x}' \mid \bm{x})P_k(\bm{x})}{P_k(\bm{x}')}.
\end{align*}
$Q_{\mrm{f}}(\bm{X})$ expresses the forward transition process from the initial to the target distribution, and $Q_{\mrm{b}}(\bm{X})$ expresses the backward process.
From equations (\ref{eqn:AIS_weight_normalized})--(\ref{eqn:backward_process}),
\begin{align}
\omega_{\mrm{norm}}(\bm{X}) = \prod_{k=1}^K \frac{P_k(\bm{x}^{(k)})}{P_{k-1}(\bm{x}^{(k)})} = \frac{Z_0}{Z(\beta)} \omega(\bm{X})
\end{align}
is obtained, where 
\begin{align*}
\omega(\bm{X}) = \exp\Big(- \beta\sum_{k=1}^K (\beta_k - \beta_{k-1})E(\bm{x}^{(k)})\Big)
\end{align*}
is the unnormalized importance weight defined in equation (\ref{eqn:AIS_weight}). 
Equation (\ref{eqn:expectation-f_AIS}) can be viewed as 
the sampling approximation of equation (\ref{eqn:expectation_f_extension}), i.e.,  
using $N$ different sequences, $\mbf{X}_{1}, \mbf{X}_{2}, \ldots, \mbf{X}_{N}$, obtained from $N$ parallel samplings from $Q_{\mrm{f}}(\bm{X})$ 
(the sampling processes shown in equation (\ref{eqn:AIS_sampling})), 
\begin{align}
\ave{f(\bm{x})}_{\beta}\approx \frac{1}{N}\sum_{\mu=1}^N\omega_{\mrm{norm}}(\mbf{X}_{\mu})f(\mbf{x}_{\mu}^{(K)})
\label{eqn:expectation-f_AIS_origin}
\end{align}
is obtained, where $\mbf{X}_{\mu} = \{\mbf{x}_{\mu}^{(k)} \in \{-1,+1\}^n\mid k = 1,2,\ldots, K\}$.
Moreover, to avoid the evaluation of the partition function, ratio $r(\beta):=Z_0 / Z(\beta)$ is approximated by $N / \Omega$ in equation (\ref{eqn:expectation-f_AIS}):
\begin{align}
1 = \sum_{\bm{X}} \omega_{\mrm{norm}}(\bm{X})Q_{\mrm{f}}(\bm{X}) &= r(\beta) \sum_{\bm{X}} \omega(\bm{X})Q_{\mrm{f}}(\bm{X})\nn
&\approx \frac{r(\beta)}{N}\sum_{\mu=1}^N\omega(\mbf{X}_{\mu}).
\label{eqn:approximation_r(beta)}
\end{align}

In the following, the asymptotic variance of the approximation of equation (\ref{eqn:expectation-f_AIS}) is considered.  
Here, the annealing schedule is assumed to be sufficiently slow, i.e., $\beta_{k} - \beta_{k-1} = \varepsilon \ll 1$. 
Based on this assumption, $\omega(\bm{X})$ and $f(\bm{x}^{(K)})$ are considered to be almost independent under $Q_{\mrm{f}}(\bm{X})$ (as well as under $Q_{\mrm{b}}(\bm{X})$) 
because the correlations between the distant variables (e.g., $\bm{x}^{(K)}$ and $\bm{x}^{(1)}$) are expected to be negligible 
(in other words, the dependency of $\omega(\bm{X})$ on $\bm{x}^{(K)}$ is expected to be negligible). 
With this assumption, the asymptotic variance is estimated as~\cite{AIS2001}
\begin{align}
V_{\mrm{AIS}}[f(\bm{x})]&\approx\frac{1}{N} W V_{\beta}[f(\bm{x})],
\label{eqn:Variance_AIS}
\end{align}
where $V_{\beta}[f(\bm{x})]$ is the variance of $f(\bm{x})$; 
here, $W\geq 1$ is the constant factor obtained from the variance of $\omega(\bm{X})$, and is independent of $f(\bm{x})$.
The factor $W$ may be close to $1$ when the target distribution has few isolated modes~\cite{AIS2001} .

\subsection*{Acknowledgment}
This work was partially supported by JSPS KAKENHI (grant Numbers 15H03699, 18K11459, and 18H03303), 
JST CREST (grant Number JPMJCR1402), and the COI Program from the JST (grant Number JPMJCE1312).

\bibliography{citation}
\end{document}